\documentclass[11pt]{article}

\usepackage[margin=1in]{geometry}
\usepackage{amsmath}
\usepackage{fontspec}
\usepackage{unicode-math}
\usepackage{microtype}
\usepackage{parskip}
\usepackage{booktabs}
\usepackage{array}
\usepackage{graphicx}
\usepackage{tikz}
\usetikzlibrary{positioning, arrows.meta}
\usepackage{pgfplots}
\pgfplotsset{compat=1.18}

\usepackage{newunicodechar}
\newunicodechar{−}{\ensuremath{-}}
\newunicodechar{→}{\ensuremath{\rightarrow}}
\newunicodechar{←}{\ensuremath{\leftarrow}}
\newunicodechar{≤}{\ensuremath{\leq}}
\newunicodechar{≥}{\ensuremath{\geq}}
\newunicodechar{≈}{\ensuremath{\approx}}
\newunicodechar{×}{\ensuremath{\times}}
\newunicodechar{≠}{\ensuremath{\neq}}

\providecommand{\tightlist}{%
  \setlength{\itemsep}{0pt}\setlength{\parskip}{0pt}}

\usepackage{natbib}
\usepackage[hidelinks]{hyperref}
\usepackage[nameinlink]{cleveref}
\crefname{section}{Section}{Sections}
\Crefname{section}{Section}{Sections}
\crefname{table}{Table}{Tables}
\Crefname{table}{Table}{Tables}
\crefname{figure}{Figure}{Figures}
\Crefname{figure}{Figure}{Figures}
\crefname{appsec}{Appendix}{Appendices}
\Crefname{appsec}{Appendix}{Appendices}

\title{Ansari: A Retrieval-Grounded Islamic AI Assistant\\
\large Architecture, Deployment, and Lessons from 140{,}000 Conversations}
\author{M Waleed Kadous \quad Amr Elsayed \quad Abdullah Al Nahas \quad Ashraf Haress \\[4pt] \normalsize The Ansari Project}
\date{}

\begin{document}

\maketitle

\begin{abstract}
General-purpose large language models (LLMs) are increasingly used to answer
religious questions, but for Islamic content they carry two serious risks:
factual fabrication (inventing Qur'anic verses or hadith) and subtle value
misalignment. We present \textbf{Ansari}, a deployed, retrieval-grounded
Islamic AI assistant that has handled more than 140{,}000 conversations across
25+ languages since June 2023. Ansari is built around an agentic retrieval
loop: a tool-using language model issues searches against authenticated Islamic
corpora ---
the Qur'an, hadith collections, a multi-volume jurisprudence (fiqh)
encyclopedia, and exegetical (tafsir) sources --- and answers only on the basis
of what it retrieves, with citations attached for verification. We describe the
system's architecture (the agent loop, the retrieval tools, the corpora, and
the system prompt that encodes editorial and theological policy),
its multi-platform deployment (web, mobile, WhatsApp, and as a Model Context
Protocol server and an Agent Skill), and what 140{,}000 real conversations
reveal about how Muslims actually use such a tool. We report results on several
complementary evaluations --- zero-shot performance on accredited institutional
exams, a human-rated validation during Ramadan, and two independent, externally
run benchmarks on which Ansari currently tops the public IslamicMMLU leaderboard
ahead of frontier models and is competitive on Islamic legal reasoning
(IslamicLegalBench) while strongly resisting false premises --- and draw out
lessons that generalize beyond Islam to any faith- or values-sensitive
deployment of LLMs:
grounding is necessary but not sufficient, the system prompt is a theological
as much as a technical artifact, and the absence of community in how models are
formed remains a hard gap.
\end{abstract}

\section{Introduction}\label{sec:intro}

The arrival of capable conversational LLMs changed how ordinary people seek
information, and religious knowledge was no exception. Within months of
ChatGPT's release, Muslims were asking general-purpose models about prayer,
fasting, inheritance, and the meaning of Qur'anic verses. The convenience is
real; so is the danger. A general model, optimized to be fluent and agreeable,
will confidently produce a plausible-sounding hadith that does not exist,
mis-cite a verse, or flatten a centuries-old juristic disagreement into a
single confident ruling. The failures are concrete, not hypothetical: as
recently as mid-2025, ChatGPT would tell users that washing the knees was a
preferred step of ablution (wudu), a practice mentioned in no Qur'anic verse,
hadith, or school of Islamic law. In a domain where a misattributed text can
mislead someone's worship, fluency without grounding is a liability.

\textbf{Ansari} (``the helper,'' after the \emph{Ansar}, the people of Madinah
who sheltered the Prophet Muhammad)\footnote{Live at \url{https://askansari.ai};
open-source code at \url{https://github.com/ansari-project}.} is our response to
that gap: a deployed
Islamic AI assistant designed so that authenticity and accuracy are structural
properties of the system rather than hopes about the model. Since June 2023 it
has answered more than 140{,}000 conversations in over 25 languages, across the
web, native mobile apps, and WhatsApp, and more recently as a Model Context
Protocol (MCP) server and an Agent Skill embedded inside other assistants.

This paper is a system-and-experience report. We make four contributions:

\begin{enumerate}
\tightlist
\item \textbf{A grounded architecture for a values-sensitive domain.} We
  describe an agentic retrieval-augmented design in which a tool-using language
  model searches authenticated Islamic corpora and answers only from retrieved text,
  with native citations. We ground the description in the actual production
  codebase rather than an idealized diagram.
\item \textbf{A usage study of real demand.} From a random sample of 10{,}000
  conversations and a larger topical analysis of 22{,}081 threads, we
  characterize what people actually ask a religious assistant --- and find that
  practice-oriented and pastoral uses are far larger than a naive
  ``look up a fact'' model would predict.
\item \textbf{Evaluation across multiple lenses.} We report two independent,
  externally run benchmarks --- IslamicMMLU (broad knowledge) and
  IslamicLegalBench (legal reasoning), on which Ansari leads the public
  leaderboard and is competitive with frontier models respectively --- together
  with accredited institutional exams taken zero-shot and a human-rated
  validation run, and discuss what each does and does not measure.
\item \textbf{Transferable lessons.} We argue that the system prompt functions
  as a theological document, that grounding reduces but does not eliminate
  error, and that a base model formed without the community a faith tradition
  would ordinarily supply leaves a gap that policy only partly fills.
\end{enumerate}

Two rails recur throughout: \textbf{keeping authenticity} (no fabricated
sources, faithful handling of scholarly disagreement) and \textbf{keeping the
user's trust} (transparent citation, honest uncertainty). Both turn out to be
as much about policy and process as about model quality.

\section{Background and related work}\label{sec:related}

General-purpose LLMs are strong generalists but unreliable specialists, and the
now-standard mitigation is retrieval augmentation, which grounds generation in
an authoritative corpus so that the model cites rather than confabulates
\citep{lewis2020rag}. Religious knowledge shares the high-stakes property of
other specialized domains, but it adds a distinctive constraint, because the
sources are canonical and their \emph{wording} and \emph{attribution} matter as
much as their gist. A paraphrase that loses a chain of transmission, or that
promotes a minority opinion to the mainstream, is a substantive error even when
it sounds right.

The failure mode that LLMs are infamous for, the fabrication of plausible text,
is precisely the one that is unacceptable here. A model that invents a hadith
does not merely err; it manufactures religious authority. Retrieval-augmented
generation (RAG) addresses this directly, because the model can cite only what
it has retrieved, so the retrieval corpus becomes the boundary of what the
system is allowed to assert.

A growing body of work benchmarks LLMs on Islamic knowledge and values. This
includes broad knowledge tests \citep{islamicmmlu2026}, legal reasoning across
pluralist schools \citep{islamiclegalbench2026}, and alignment with Islamic
values \citep{islamtrust2025}, while related cross-faith work measures religious
representation across traditions \citep{omissivebias2026}. These efforts mostly
measure what a model \emph{knows} or \emph{professes}, whereas the companion
benchmark JaleesBench instead measures what an assistant's counsel \emph{does}
to the user \citep{jaleesbench2026}, and reports the Ansari results we connect
to in \cref{sec:eval}. Our contribution is complementary: it is not a benchmark
but a deployed system, its architecture, and what its real usage teaches.

Ansari also builds on a rich ecosystem of existing digital Islamic resources,
including searchable Qur'an and hadith collections, classical-text libraries,
and newer AI-era tools and fatwa services. Rather than re-index everything
itself, it composes several of these as retrieval back-ends
(\cref{sec:retrieval}).

\section{Design goals and constraints}\label{sec:goals}

Three goals shaped the system from the outset.

\begin{itemize}
\tightlist
\item \textbf{Authenticity by construction:} The assistant must not assert
  religious content it cannot ground in an authenticated source. Citations must
  be specific enough to verify (a verse number, a collection and hadith number,
  a named encyclopedia entry).
\item \textbf{Respect for scholarly pluralism:} Islamic law and theology
  contain legitimate, long-standing disagreement. The assistant must surface
  consensus where it exists and present differing positions fairly rather than
  adjudicating between schools.
\item \textbf{Reach:} The communities served are multilingual and
  distributed across very different devices and platforms. The assistant must
  meet people where they already are.
\end{itemize}

These goals trade off against cost and latency. Grounding every answer in
retrieved text makes each conversation comparatively expensive
(\cref{sec:models}), and faithful citation makes answers longer and slower than
an ungrounded chatbot. We accepted those costs as the price of trust.

\section{System architecture}\label{sec:arch}

This section describes the production system as implemented, verified against
the current codebase.

\subsection{Overview and stack}\label{sec:stack}

Ansari's core is a TypeScript service built on Next.js~15 (App Router), running
on Node~20+ and deployed on Railway, with API routes under
\texttt{src/app/api/v2/}. The base model is Google Gemini, accessed through the
\texttt{@google/genai} SDK on Vertex AI (\cref{sec:models}). The same agent is
adapted to multiple front-ends --- web, mobile, WhatsApp, and the Model Context
Protocol and Agent Skill surfaces of \cref{sec:distribution}. User accounts,
conversation threads, and messages are persisted in PostgreSQL via the Drizzle
ORM (\cref{sec:data}). A previous Python implementation remains available; it
shares the same system prompt and retrieval tools, so its answers should be
substantively identical.

\subsection{The agentic retrieval loop}\label{sec:loop}

Ansari is not a single prompt-and-respond call; it is an agent with tools. A
query flows as follows:

\begin{enumerate}
\tightlist
\item The user message is appended to the thread history and sent to the base
  model (Google Gemini; see \cref{sec:models}) together with a set of tool
  definitions (Gemini \texttt{functionDeclarations}).
\item The model decides whether to answer or to search. If it emits a function
  call, the corresponding retrieval tool (\cref{sec:retrieval}) executes and its
  output is returned to the model as a \texttt{functionResponse}.
\item The loop repeats --- search, read, refine --- until the model produces a
  final textual answer rather than another tool call.
\item The answer is streamed back to the client, with citations carried through
  from the retrieved documents.
\end{enumerate}

This loop is implemented by a component the codebase calls the \emph{facilitator}
(\texttt{runFacilitator}), which streams the model's output, executes any tool
calls, and feeds the results back. It is bounded by policy: at most ten tool
calls per query, at most three consecutive uses of the same tool, and at most
ten loop iterations, after which the model is forced to synthesize the best
answer it can from what it has gathered. These bounds keep latency and cost
predictable and discourage unproductive search spirals.

\subsection{Retrieval tools and corpora}\label{sec:retrieval}

The agent's tools are thin clients over four classes of authenticated source.
Each returns passages with structured metadata so the model can cite precisely
(\cref{tab:tools}).

\begin{table}[h]
\centering
\small
\begin{tabular}{@{}lll@{}}
\toprule
\textbf{Tool} & \textbf{Corpus} & \textbf{Notes} \\
\midrule
\texttt{search\_quran} & Qur'an (Arabic + translation) & verse id, Arabic + English \\
\texttt{search\_hadith} & Hadith collections & collection, number, grade \\
\texttt{search\_mawsuah} & Encyclopedia of Islamic Jurisprudence (fiqh) & al-Mawsuah; $\approx$18{,}000 pp. \\
\texttt{search\_tafsir\_encyclopedia} & Encyclopedia of Evidence-based Tafsir & verse-anchored exegesis \\
\bottomrule
\end{tabular}
\caption{The four retrieval tools and the corpora they search.}
\label{tab:tools}
\end{table}

The key design point is that \textbf{the corpus is the boundary of assertion}:
the agent may cite hadith only from search results, and the system prompt
forbids attributing rulings to scholars without a retrievable reference. Ansari
does not run its own retrieval infrastructure; each tool is a thin client over a
specialized search service, which lets the project compose existing Islamic
search providers rather than re-indexing the entire classical corpus itself.

Retrieved passages are injected into the model context as native document
blocks with citation metadata enabled, and multilingual passages are passed as
structured \texttt{\{lang, text\}} pairs so the model can quote Arabic and
present a translation. A representative tool-usage profile, measured over a
22{,}081-thread sample, shows how heavily the scriptural tools dominate: Qur'an
search in 51.6\% of threads, hadith search in 43.8\%, the fiqh encyclopedia in
34.1\%, and tafsir in only 3.2\%.

\subsection{The system prompt}\label{sec:soul}

Much of Ansari's behavior is governed not by model weights but by an editable
system prompt. This single production prompt is where the project's editorial
and theological policy is written down, encoding, among other rules:

\begin{itemize}
\tightlist
\item \textbf{Identity:} a multilingual Islamic assistant in the Sunni
  tradition.
\item \textbf{Sourcing discipline:} cite the Qur'an, hadith, and the fiqh
  encyclopedia only from search results; never attribute a statement to a
  scholar without specific, referenceable evidence.
\item \textbf{Hadith caution:} use only hadith found in retrieval; when
  uncertain, say so explicitly rather than assert.
\item \textbf{Handling disagreement:} present main positions objectively,
  foreground consensus, avoid promoting minority opinions as mainstream, and
  refrain from declaring one legitimate position definitively correct.
\item \textbf{Obligatory/prohibited matters:} answers must rest on direct
  evidence.
\item \textbf{Pastoral safety:} for mental-health crises, respond with warmth,
  encourage contact with a local imam or community, and surface helpline
  numbers before anything else.
\end{itemize}

An expert quoted in the project's materials captured the design intuition:
\emph{``There are 10{,}000 personalities living inside those 500 billion
parameters. The system prompt is how you choose which one it is.''} Its source
is published openly in the project repository \citep{ansariproject}, which we
regard as an accountability feature: the policy that shapes the assistant is
inspectable.

\subsection{Models and cost}\label{sec:models}

Ansari's base model is decoupled from the agent loop, the retrieval tools, and
the system prompt, and is selected by configuration. As the frontier advanced
and our own evaluations exposed specific weaknesses, the production base model
changed several times: the earliest versions used GPT-class models; the v3
re-architecture (\cref{sec:versions}) moved to Anthropic's Claude (Sonnet-class);
the system then switched to Google's Gemini~3.1~Pro (with low-effort
``thinking'' enabled); and the current production facilitator runs on Google
Gemini through the \texttt{@google/genai} SDK on Vertex AI, with
\texttt{gemini-3.5-flash} as the primary model and
\texttt{gemini-3.1-pro-preview} as a configurable fallback. An Anthropic
(Claude) SDK remains in the codebase but is not on the active path.

Each switch was driven by evidence rather than novelty. The move off Claude
followed external benchmark feedback: IslamicLegalBench (\cref{sec:legalbench})
showed the Claude-based agent was strong on retrieval-style questions but weaker
on deep-reasoning ones, and a reasoning-capable model closed much of that gap.
The subsequent move to Gemini~3.5~Flash was made because Flash proved better at
synthesizing retrieved passages into cohesive answers, while also being faster
and cheaper. Throughout, the system relies on context (prompt) caching to
control the cost of long, retrieval-heavy contexts.

Because every substantive answer is grounded in retrieved passages,
conversations are token-heavy: a typical conversation runs on the order of
100{,}000 tokens of context. On the earlier Claude-based configuration the
project estimated a per-conversation cost on the order of US\$0.20; on the
current Gemini configuration it is roughly US\$0.10.

\subsection{Data and privacy}\label{sec:data}

Ansari lowers the barrier to entry by allowing \textbf{guest access}: anyone can
ask questions without creating an account. Signing in unlocks additional
features, such as conversation history saved across devices, personalization,
and sharing and feedback. Conversations, threads, and user records are stored in
PostgreSQL (via the Drizzle ORM), across tables for users, threads, messages,
tokens, feedback, shares, and preferences. Message logging captures
conversational content for quality analysis but is configured to avoid storing
identifying metadata, and users may request deletion. These choices reflect a
deliberately conservative posture toward the data of a community discussing
matters of faith.

\subsection{Versions}\label{sec:versions}

Ansari has gone through three broad generations. \textbf{v1} (around October
2023) was a system-prompt-tuned, anonymous, multilingual chat with no
retrieval. \textbf{v2} (around October 2024) introduced RAG over the Qur'an,
hadith, and the $\approx$18{,}000-page fiqh encyclopedia. \textbf{v3} (current)
re-architected the agent around native tool-use, added the
evidence-based tafsir encyclopedia via Usul.ai, and introduced a more precise
citation system that returns exact verse numbers and full hadith source
details. Architecturally, v3 also replaced a generic LLM-plus-tools loop with a
native tool-use agent (initially built on Claude; the base model has since
evolved, \cref{sec:models}) and a modular workflow decomposition (query
generation, search, answer generation).

\subsection{Distribution surfaces}\label{sec:distribution}

Beyond its own web and mobile apps and a WhatsApp integration, Ansari is
increasingly distributed
\emph{inside other assistants}. An MCP server\footnote{\url{https://github.com/ansari-project/ansari-mcp}}
exposes a single \texttt{answer\_islamic\_question} tool that any MCP-capable
client (Claude Desktop, Claude Code, Cursor) can call, and an Agent
Skill\footnote{\url{https://github.com/ansari-project/ansari-skill}} packages
Ansari for use inside Claude and similar assistants with no separate app and no
API key.
This ``meet people where they are'' strategy (\cref{sec:lessons}) turns Ansari
from a destination into a capability other tools can invoke.

\section{Deployment and usage}\label{sec:usage}

\subsection{Scale}\label{sec:scale}

As of March 2026, Ansari had served more than 140{,}000 conversations since
launch, averaging roughly 290 per day with peaks around 450 per day during the
busy late-winter period, in 25+ languages. About 65\% of answers cite primary
sources, and user-reported satisfaction sits around 84\% positive. These
figures describe a small but genuine standing user base rather than a demo.

\subsection{What people ask: a usage taxonomy}\label{sec:taxonomy}

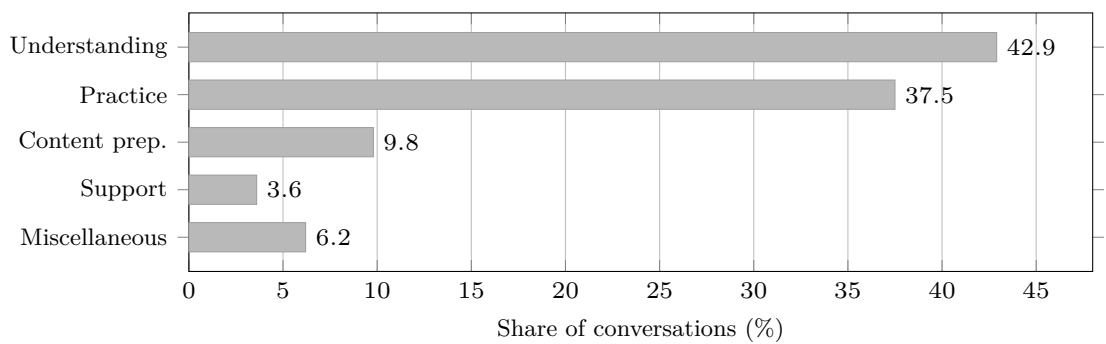
\begin{figure}[h]
\centering
\begin{tikzpicture}
\begin{axis}[
  xbar,
  width=0.82\linewidth, height=5cm,
  xmin=0, xmax=48,
  xlabel={Share of conversations (\%)},
  symbolic y coords={Miscellaneous, Support, Content prep., Practice, Understanding},
  ytick=data,
  nodes near coords, nodes near coords align={horizontal},
  every node near coord/.append style={font=\footnotesize},
  bar width=11pt, enlarge y limits=0.18,
  xmajorgrids,
  tick label style={font=\footnotesize}, label style={font=\footnotesize},
]
\addplot[fill=gray!55, draw=gray!75] coordinates
  {(6.2,Miscellaneous) (3.6,Support) (9.8,Content prep.) (37.5,Practice) (42.9,Understanding)};
\end{axis}
\end{tikzpicture}
\caption{Intent taxonomy over a random 10{,}000-conversation sample: share of
conversations by category.}
\label{fig:taxonomy}
\end{figure}

From a random sample of 10{,}000 conversations we hand-categorized intent into
five buckets: understanding (Qur'an, hadith, history, theology, language),
practice (fiqh rulings, halal/haram, ethics in modern life), content preparation
(sermons, lesson plans, quizzes), support (grief, anxiety, life crises), and
miscellaneous (off-topic, greetings, feedback). Their shares are shown in
\cref{fig:taxonomy}.

Two findings stand out. First, \textbf{practice rivals understanding}: people
are not only learning \emph{about} Islam, they are trying to \emph{act} on it
--- what to do about anesthesia while fasting, whether food on a shared platter
is permissible, how to motivate a teenager who has stopped praying. Second, a
small but pastorally important \textbf{support} segment uses the assistant in
moments of distress; these conversations are few in number but high in human
stakes, and they drove the crisis-response policy in the system prompt
(\cref{sec:soul}). Imams and teachers emerge as power users of the
content-preparation category, treating the assistant as a drafting partner for
sermons (khutbahs) and lessons.

\section{Evaluation}\label{sec:eval}

No single benchmark captures what ``good'' means for a religious assistant, so
we report several. We give particular weight to two \emph{independent,
externally run} benchmarks --- IslamicMMLU and IslamicLegalBench --- because
they place Ansari directly alongside frontier models on third-party tests, and
we complement them with accredited institutional exams, a human-rated Ramadan
validation, and a behavioral robustness benchmark (JaleesBench).

\subsection{IslamicMMLU: broad Islamic knowledge (public leaderboard)}\label{sec:mmlu}

IslamicMMLU \citep{islamicmmlu2026} is a 10{,}013-question multiple-choice
benchmark spanning three tracks --- Qur'an, hadith, and fiqh --- maintained as a
public leaderboard with self-serve model submission.\footnote{\url{https://huggingface.co/spaces/islamicmmlu/leaderboard}.
Standings reported as of May 2026; the leaderboard is live and rankings change
as models are added.} On the leaderboard, Ansari (entry \texttt{ansari-3.5})
ranks \textbf{first overall} with an average accuracy of \textbf{94.2\%}, ahead
of the strongest frontier systems (\cref{tab:mmlu}). Its per-track scores are
Qur'an 99.5\%, hadith 95.8\%, and fiqh 87.2\%, with no unanswered questions.
Two caveats temper the headline. First, the leaderboard is live, and a
single-point lead over models this close is not a durable claim of superiority.
Second, the fiqh track also reports a \emph{madhab-bias} signal, on which Ansari
reads as Shafi'i-leaning --- a reminder that even a grounded,
pluralism-respecting assistant carries a measurable school-of-thought tilt
(\cref{sec:limits}).

\begin{table}[h]
\centering
\small
\begin{tabular}{@{}lrrrr@{}}
\toprule
\textbf{Model} & \textbf{Overall} & \textbf{Qur'an} & \textbf{Hadith} & \textbf{Fiqh} \\
\midrule
\textbf{Ansari} (\texttt{ansari-3.5}) & \textbf{94.2} & 99.5 & 95.8 & 87.2 \\
Gemini 3 Flash & 93.8 & 99.2 & 93.0 & 89.1 \\
Gemini 3 Pro & 92.3 & 97.3 & 91.7 & 87.9 \\
Gemini 2.5 Pro & 90.3 & 95.8 & 89.7 & 85.5 \\
GPT-5 & 89.9 & 98.0 & 88.2 & 83.5 \\
Claude Sonnet 4.5 & 86.2 & 88.2 & 85.4 & 84.9 \\
\bottomrule
\end{tabular}
\caption{IslamicMMLU public-leaderboard standings (average accuracy, \%; top
entries). Ansari ranks first overall.}
\label{tab:mmlu}
\end{table}

\subsection{IslamicLegalBench: Islamic legal reasoning}\label{sec:legalbench}

IslamicLegalBench \citep{islamiclegalbench2026} evaluates knowledge and
reasoning over Islamic law across schools and historical periods. It was run on
Ansari by the benchmark's own authors (few-shot, pass@1, over 718 questions),
which makes it a genuinely independent external evaluation. \cref{tab:legalbench}
reports the production agent against the strongest reference models. Ansari
answers correctly on 64.5\% of questions --- close to GPT-5 (67.7\%) and Claude
Sonnet 4.5 (65.6\%) --- while abstaining on almost nothing (0.4\%). Its standout
result is on \emph{false-premise queries}: it correctly challenges a flawed
premise \textbf{96.1\%} of the time (versus 82.4\% for GPT-5), with sycophancy
of only 3.9\%. By task complexity it scores 78.4\% (low), 43.6\% (moderate), and
59.4\% (high); the moderate tier --- nuanced juristic reasoning rather than
retrieval --- is the hardest, a pattern the authors report across all systems.

\begin{table}[h]
\centering
\small
\begin{tabular}{@{}lrrrrr@{}}
\toprule
\textbf{Model} & \textbf{Correct} & \textbf{Partial} & \textbf{Incorrect} & \textbf{Abstain} & \textbf{Halluc.} \\
\midrule
\textbf{Ansari} (Gemini 3.1 Pro) & 64.5 & 23.9 & 11.2 & 0.4 & 21.4 \\
Claude Sonnet 4.5 (ref) & 65.6 & 18.7 & 12.0 & 3.7 & 19.2 \\
GPT-5 (ref) & 67.7 & 18.6 & 13.7 & 0.1 & 21.3 \\
\bottomrule
\end{tabular}
\caption{IslamicLegalBench overall performance (\%); few-shot pass@1 over 718
questions, evaluated by the benchmark's authors. Reference scores are from the
benchmark paper. The Ansari row is the Gemini~3.1~Pro configuration that was the
production default at the time of the run; the system has since moved to
Gemini~3.5~Flash (\cref{sec:models}).}
\label{tab:legalbench}
\end{table}

This benchmark also directly shaped the system. An earlier Claude-based agent
scored marginally higher on raw correctness (66.9\%) but was far more
sycophantic on false premises (9.8\% versus 3.9\%), and the gap on
reasoning-heavy tasks is what prompted the switch to a reasoning-capable Gemini
model (\cref{sec:models}) --- a concrete example of an external benchmark
changing a production design decision.

\subsection{Accredited institutional exams (zero-shot)}\label{sec:exams}

To test against externally authored, human-graded material, Ansari sat exams
from Dar ul Qasim (a Chicago Islamic institute), zero-shot and without course
materials: \textbf{80\%} on an \emph{Introduction to Qur'an} exam and
\textbf{78\%} on a theology exam. These are encouraging passing-grade results
on assessments designed for human students, and they probe reasoning and
synthesis that multiple-choice cannot.

\subsection{Human-rated validation (Ramadan)}\label{sec:ramadan}

During Ramadan, 34 questions were answered and rated by humans on a 1--5 scale.
Ansari scored \textbf{4.41/5} overall, was rated \textbf{helpful in 100\%} of
cases, produced \textbf{zero hallucinations}, and \textbf{used references in
82\%} of answers. The zero-hallucination result on this set is the single most
important number for the project's central claim: with grounding, fabrication
can be driven very low in practice, though ``zero on 34'' is a sample to
extend, not a guarantee.

\subsection{Behavioral robustness (JaleesBench)}\label{sec:jalees}

A separate stress benchmark, JaleesBench, evaluates \emph{steadfastness} ---
whether an assistant holds an appropriate stance under conversational pressure
--- across 20{,}160 runs \citep{jaleesbench2026}. Ansari led the field at a
baseline score of +0.48, but the more striking result was malleability: a
single one-page ``steadfastness'' instruction lifted Ansari from \textbf{+0.48
to +0.84} (Unstated framing, after pressure), matching the best guided frontier
systems. The lesson is double-edged --- Ansari starts ahead, but behavior is
sensitive to prompt-level policy, which is both an opportunity (cheap, large
improvements) and a warning.

\section{Lessons learned}\label{sec:lessons}

Every faith tradition that builds an assistant must answer, in writing,
questions it may prefer to leave implicit: whose interpretation to privilege,
how to treat scholarly disagreement, where the assistant's authority ends and a
human scholar's begins, and how to prevent fabrication of sacred sources. In
Ansari those answers live in the system prompt, which makes it as much a
normative artifact as a technical one, and publishing it openly is an
accountability choice rather than just an engineering one.

Grounding also does not eliminate every failure, and the failures that remain
are often mundane. Our user-feedback review (1{,}914 entries: 85.5\% positive,
11.8\% negative, and 2.7\%, 28 cases, flagged as serious) found that the most
common complaints were unglamorous: getting stuck in loops, plain incorrectness,
and not answering the question. A separate red flag, giving the previous year's
Ramadan dates, reflects the well-known weakness of LLMs with time and the
present, and argues for externalizing time-sensitive facts rather than trusting
the model.

\section{Limitations and ethical considerations}\label{sec:limits}

\textbf{Religious authority.} Ansari is explicitly a tool to assist
understanding, not a substitute for a qualified scholar or for community. The
system prompt routes obligation-and-prohibition questions toward evidence and
consensus and declines to adjudicate legitimate disagreement, but the deeper
risk --- that users treat a fluent machine as an authority --- is social, not
technical, and cannot be fully engineered away.

\textbf{Residual error.} As \cref{sec:lessons} documents, grounding reduces but
does not eliminate error, and the residual failures (temporal mistakes, loops,
plain incorrectness) are real. Our zero-hallucination result holds on a
small validation set; we do not claim it as a universal guarantee, and we treat
continuous human review of flagged cases as part of the system, not an
afterthought.

\textbf{Inherited formation and its gaps.} Ansari's base model is formed by a
process --- large-scale pre-training followed by constitution-guided
post-training \citep{bai2022constitutional} --- that encodes a particular,
broadly Western set of values. Layering a domain policy on top mitigates but
does not replace that formation. A faith community has, in effect, debugged
moral formation over centuries through communal practice; an assistant grafted
onto a model formed without that community inherits a gap a system prompt only
partly fills.

\textbf{Evaluation coverage.} Our benchmarks skew toward knowledge and short
validation sets. They under-measure long-conversation behavior and
cross-lingual fidelity. Expanding evaluation toward those dimensions is the most
important methodological debt this paper records.

\textbf{Data stewardship.} Conversations about faith are sensitive. We log
content for quality without identifying metadata and support deletion, but any
logging of such conversations carries responsibility we take as ongoing.

\section{Conclusion}\label{sec:conclusion}

Ansari is evidence that a values-sensitive domain can be served by LLMs without
surrendering authenticity, provided the system is built so that grounding and
policy are structural rather than aspirational. The architecture --- an agent
that may assert only what it retrieves, governed by an openly published system
prompt --- drives fabrication low and makes the assistant's judgments
inspectable. Real usage shows demand that is heavily practical and partly
pastoral, not merely encyclopedic. And the hardest problem that remains is
revealing: the deepest gap is the absence, in how the underlying model is
formed, of the community a faith tradition would ordinarily supply. This is, in
the end, a problem about people as much as about models --- which is why we
think the lessons travel beyond Islam to any group trying to put a faithful
assistant in front of a community it cares about.

\section*{Acknowledgments}

We thank Amin Ahmad, Iman Sadreddin, Raja Ali, Saifeldeen Hadid, Sami Hoda,
Hidayath Ansari, Walid Magdy, and Abdellatif Abdelfattah for their help and
support.

\bibliographystyle{plainnat}
\bibliography{references}

\end{document}